\documentclass[conference]{IEEEtran}
\IEEEoverridecommandlockouts
\usepackage{cite}
\usepackage{amsmath,amssymb,amsfonts}
\usepackage{mathptmx}
\usepackage{float}
\usepackage{graphicx}
\usepackage{textcomp}
\usepackage{xcolor}
\usepackage{caption}
\usepackage{subfigure}
\usepackage{ntheorem}
\usepackage{xcolor}
\usepackage{stfloats}
\usepackage[ruled,linesnumbered,vlined]{algorithm2e} 
\usepackage[]{hyperref}
\usepackage{comment}
\usepackage{overpic}

\def\BibTeX{{\rm B\kern-.05em{\sc i\kern-.025em b}\kern-.08em
    T\kern-.1667em\lower.7ex\hbox{E}\kern-.125emX}}
\begin{document}

\newcommand{\bG}{\bar{G}}
\newcommand{\bV}{\bar{V}}
\newcommand{\bE}{\bar{E}}
\newcommand{\vE}{\vec{E}}
\newcommand{\vG}{\vec{G}}
\newcommand{\NN}{\mathbb{N}}
\newcommand{\cH}{\mathcal{H}}
\newcommand{\cV}{\mathcal{V}}
\theorembodyfont{\rmfamily}
\newtheorem{assum}{\textbf{Assumption}}  
\newtheorem{theorem}{\textbf{Theorem}}
\newtheorem{definition}{\textbf{Definition}}
\newtheorem{lemma}{\textbf{Lemma}}
\newtheorem{remark}{\textbf{Remark}}
\newtheorem{proposition}{\textbf{Proposition}}

\newtheorem{problem}{\textbf{Problem}}

\newtheorem*{Proof}{Proof}
\newtheorem{Corollary}{Corollary}
\newcommand{\bP}{\bar{P}}
\newcommand{\bp}{\bar{p}}
\newcommand{\bX}{\bar{X}}
\newcommand{\bx}{\bar{x}}
\newcommand{\mE}{\mathcal{E}}
\newcommand{\bmE}{\bar{\mathcal{E}}}

\title{Integer-Programming-Based Narrow-Passage Multi-Robot Path Planning with Effective Heuristics
}


\author{Jiaxi Huo, Ronghao Zheng, Meiqin Liu, Senlin Zhang
\thanks{All the authors are with the College of Electrical Engineering, Zhejiang University, Hangzhou, 310027, China
(Email: jiaxi.huo@zju.edu.cn, rzheng@zju.edu.cn, liumeiqin@zju.edu.cn, slzhang@zju.edu.cn)}
}

\maketitle

\makeatletter
\newcommand{\removelatexerror}{\let\@latex@error\@gobble}
\makeatother

\begin{abstract}
We study optimal Multi-robot Path Planning (MPP) on graphs, in order to improve the efficiency of multi-robot system (MRS) in the warehouse-like environment. We propose a novel algorithm, OMRPP (One-way Multi-robot Path Planning) based on Integer programming (IP) method. 
We focus on reducing the cost caused by a set of robots moving from their initial configuration to goal configuration in the warehouse-like environment. The novelty of this work includes: 
(1) proposing a topological map extraction based on the property of warehouse-like environment to reduce the scale of constructed IP model; 
(2) proposing one-way passage constraint to prevent the robots from having unsolvable collisions in the passage. 
(3) developing a heuristic architecture that IP model can always have feasible initial solution to ensure its solvability. 
Numerous simulations demonstrate  the efficiency and performance of the proposed algorithm.


\end{abstract}

\begin{IEEEkeywords}
    Path Planning for Multiple Mobile Robots or Agents, Multi-Robot Systems, Collision Avoidance, Logistics.
\end{IEEEkeywords}

\section{Introduction}
We study the Multi-robot Path Planning (MPP) problem which has been explored for decades, focusing on developing a high efficient algorithm with heuristics. In this work, the robots are uniquely labeled (i.e., distinguishable). In our distinguishing narrow-passage warehouse-like environment setting, the majority of vertices in the map are occupied by shelves. Passages between shelves are narrow, i.e., the width of passages is one free vertex. Here, by narrow-passages, we means that each two robots cannot move side by side, as shown in Figure.~\ref{warehouse}. Robots should move from its current vertex to an adjacent free one in one time step without robot collision. Robot collision occurs when two robots simultaneously move to the same free vertex or along the same edge in opposing directions. Here, we study two common optimality objectives focusing on distance optimality. Distance optimality objectives have been proven to be NP-hard \cite{yu2015intractability}.

In this work, we propose the complete \textbf{OMRPP} algorithm and effective heuristics in order to generate collision-free robot path set efficiently in the narrow-passage warehouse-like environment. Applying the same optimality ratio definition in \cite{han2020ddm}, collision-free path generated by \textbf{OMRPP} has 1.x optimality ratio with efficient running time performance.

\textbf{Contributions.} This study brings two main contributions:
\begin{enumerate}
    \item Based on narrow-passage warehouse-like environment, we propose an IP-based multi-robot path planning framework. The pipeline of this framework can be applied effectively to high-density environment and produces anytime-feasible solutions.
    \item We introduce a heuristic method to boost the effectiveness of the proposed IP-based framework. In polynomial running time complexity, the solutions of the heuristic method can be achieved within performance boundary. 
\end{enumerate}

\section{Related works}

Nowadays Multi-Robot Systems (MRS) have found many applications in the real environment, e.g., logistics warehouses \cite{wurman2008coordinating,han2019effective}, in order to improve the scheduling efficiency. MRS application studies can also be found in pickup and delivery \cite{liu2019task}, localization \cite{fox2000probabilistic}, communication and coordination \cite{luna2010network}. Path planning based on MRS (briefly MPP) is an optimization problem which has been explored for decades \cite{erdem2013general,erdmann1987multiple,lavalle1998optimal,fox2000probabilistic,schouwenaars2001mixed}. MPP studies can be found its initial mathematical representation on the classical 15-puzzle moving problem \cite{johnson1879notes,loyd1959mathematical,archer1999modern}, studies have found its complexity in \cite{goldreich2011finding}. 
In this work, we focus MPP problems on the stable environment where the shelves are fixed. Readers can refer to \cite{wurman2008coordinating,van2008reciprocal} for dynamic environment setting, e.g., moving obstacles and dynamic task assignments. 

Multiple algorithms have been proposed to solve MPP problems. MPP solvers like Conflict-Based Search (CBS) \cite{sharon2015conflict} and Exact Mixed Integer Programming (MIP) \cite{schouwenaars2001mixed} can be used to generate paths set for MRS optimally. Optimal solvers (e.g., CBS and MIP) generate optimal solutions at the cost of running time. In order to improve solving efficiency, polynomial-time algorithms are proposed like Priority Inheritance with Backtracking
(PIBT) \cite{okumura2019priority}, push-and-swap-and-rotate \cite{de2013push}. Compared with optimal algorithms, these polynomial algorithms solve MPP problems fast while optimality is lost. In order to balance optimality and running time, sub-optimal solvers like Integer Linear Programming with splits \cite{yu2016optimal,han2019integer} and Enhanced CBS (ECBS) \cite{barer2014suboptimal} are designed for efficient solving process. 

In our environment setting, narrow passages impose challenges to the MPP algorithms. Partial splitting heuristic methods in \cite{han2019integer} may lose efficiency as possible alternative vertices may be excluded by splitting heuristic methods when treating robot collisions in narrow passages. Here, as we note in the main context, robot collisions in narrow passages are avoided by assigning directions to passages, similar ideas can be seen in \cite{wang2008fast,cohen2016improved}. To the best of our knowledge, we are the first to apply optimized passage directions in MPP problems rather than fixed \cite{wang2008fast} or heuristic weighted passage directions \cite{cohen2016improved}.

\section{Preliminaries}

\subsection{Multi-robot path planning in narrow-passage environment}
Let $G=(V,E)$ be an undirected grid map, with  $V := \{(i,j)| M(i,j) = 0\} $ being the vertex set. For each $(i,j) \in V$, its neighborhood is 
$N\big((i,j)\big):=\{(i+1,j),(i,j+1),(i-1,j),(i,j-1)\}\cap V$ using \emph{4-way neighbourhood}.
Here, $E:=\{(v_i,v_j)| v_i \in V \wedge v_j\in N(v_i)\}$ is defined as the edge set.
In this work, we suppose that a bijection exists between $V$ and all the grid cells represented by $M$. Let $R=\{r_1,\dots,r_n\}$ be a set of $n$ robots. 
A \emph{configuration} of the robots is an injective map from $R$ to $V$, i.e., any two robots $r_i$ and $r_j$ $(i\neq j)$ occupy different vertices of $V$. At any given time step $t\in \{0,1,\dots,\}$, the robots assume a configuration. 
The initial configurations and goal configurations of the robots are denoted as $X_I$ and $X_G$.

For a robot $r \in R$ with initial and goal configurations $X^I(r),\, X^G(r) \in V$, a scheduled path is defined as a sequence of vertices $P_r := \left\{p_r\left(t_r^0\right),
p_r\left(t_r^0+1\right),\dots,
p_r\left(t_r^0 + {T_r}\right)\right\}$, $p_r(t) \in V $ satisfying:
\begin{enumerate}
    \item $p_r \left(t_r^0\right)=X^I(r)$;
    \item $p_r\left(t_r^0+{T_r}\right)=X^G(r)$;
    \item $\forall t, t_r^0+1\leq t \leq t_r^0+T_r$, $p_r(t)\in \{p_r(t-1)\}\cup N\left(p_r(t-1)\right)$.
\end{enumerate}
Between time step $t-1$ and $t$, robot $r$ occupies an edge
$\big(p_r(t-1),p_r(t)\big) \in E$. The scheduled path set of $R$ is $P:=\{P_1,\dots,P_n\}$. 
We say that two paths $P_i,P_j$ are in collision if there exists $t\in \mathbb{Z}^+$ such that $p_i(t)=p_j(t)$ (meet collision) or $\left(p_i(t-1),p_i(t)\right) = \left(p_j(t),p_j(t-1)\right)$ (head-on collision). A path $P_i$ is \emph{feasible} if no other $P_j$ is in collison with $P_i$.

Fig.~\ref{warehouse} shows an example with the green blocs A/D 
show a portion of a collision-free path, while the red blocs C/E show head-to-head collisions and the yellow bloc B shows a crossing collision.

\begin{problem}
\textbf{(Original MPP problem)}: Given a 3-tuple $\langle G,X^I,X^G\rangle$, find a path set $P=\{P_1,\dots,P_n\}$ such that for all $1\leq i\leq n$ satisfying: $P_i$ is feasible.\label{OriginPro}
\end{problem} 

In a warehouse, a robot starts from a occupied cell carrying the shelf. 
Once leaving the occupied cell, it enters a cell in a narrow passage,
which is its initial configuration $X^I(r)$. When robot $r$ arrives at $X^G(r)$, it will go under a nearby shelf and disappear from the passages.
In such scenarios, we present the following assumption:

\begin{assum}\label{assum1}
A robot $r$ occupies a vertex of $G$ only during $t_r^0$ to $t_r^0+ T_r$, i.e., robot $r$ is not an obstacle for other robots 
for all $t, t \notin \{t_r^0,t_r^0+1,\dots,t_r^0+T_r\}$.
\end{assum}

In most of the existing MPP algorithms, 
the time step, i.e., $t_r^0, t_r^0+1, \dots, t_r^0+T_r$ 
are also an important part of the planning results of Problem~\ref{OriginPro}.
That is, robot $r$ should arrive at $p_r(t)$ at the time step $t$ exactly.
However, in real scenarios, it is very difficult to require the robots follow the scheduled paths in step consistently.
For example, the original planning result requires that each robot should move one grid cell at each time step. However, in the real scenarios, the robots may move at different velocities due to, e.g., the different weights of their loads. In such case, even only one of the robots cannot follow its scheduled path in step consistently, collisions may happen in a narrow passage. To avoid collisions in such case, 
all the robots should take the scheduled motions strictly. 
This issue motivates us to consider a variant of Problem~\ref{OriginPro} as follows.
Let us introduce the following concept first. 
\begin{definition}
An \emph{anytime-feasible path set} $P$ is define as:
\begin{align*}
	P_r =\{p_r(t_r^0), p_r(t_r^0+1), \dots, p_r(t_r^0+T_r)\},\\p_r(t_r^0)=X^I(r),p_r(t_r^0+T_r)=X^G(r)
\end{align*} 
for all $r \in \{1,2\dots,R\}$ such that for any $i,j \in \{1,2\dots,R\}$, paths $P_i$ and $P_j$ are feasible for any $t_i^0 \in \NN,T_i \in \NN, t_j^0 \in \NN$ and $T_j \in \NN$.
\end{definition} 
For an {anytime-feasible path set}, each robot only needs to follow its resulting path at any time steps.

\begin{problem}
\label{pb}
\textbf{(Anytime feasible MPP problem)}: Given a 3-tuple $\langle G,X^I,X^G\rangle$, find a path set $P=\{P_1,\dots,P_n\}$ such that for all $1\leq i\leq n$ satisfying: $P_i$ is \emph{anytime-feasible}.
\end{problem}



\subsection{Optimal Formulation}

Let $P=\{P_1,\dots,P_n\}$ be an feasible solution to some fixed MPP instance. For a path $P_i\in P$, let $len(P_i)$  denotes the length of the path $P_i$, which is the total number of times the robot $r_i$ changes its residing vertex while following $P_i$. A robot which follows a path $P_i$ may visit the same vertex multiple times. In this work, we examine two common objectives focusing on distance optimality and try to improve the time optimality under the fact that distance optimality and time optimality can not always be globally optimized simultaneously \cite{yu2015intractability}.

\emph{Objective 1: \textbf{TDMPP}} \textbf{(Total Distance)}: Compute a anytime-feasible path set $P$ that minimizes $\sum_{i=1}^nlen(P_i)$.

\emph{Objective 2: \textbf{MDMPP}} \textbf{(Maximum Distance)}: Compute a anytime-feasible path set $P$ that minimizes $\max_{1\leq i \leq n}len(P_i)$.

Here, we note that if the same vertex appears twice consecutively in $P_i$, $len(P_i)$ only accumulates once.
\subsection{Amazon warehouse-like environment representation}
\label{env}
\begin{figure}
    \centering
    \includegraphics[width=1\linewidth]{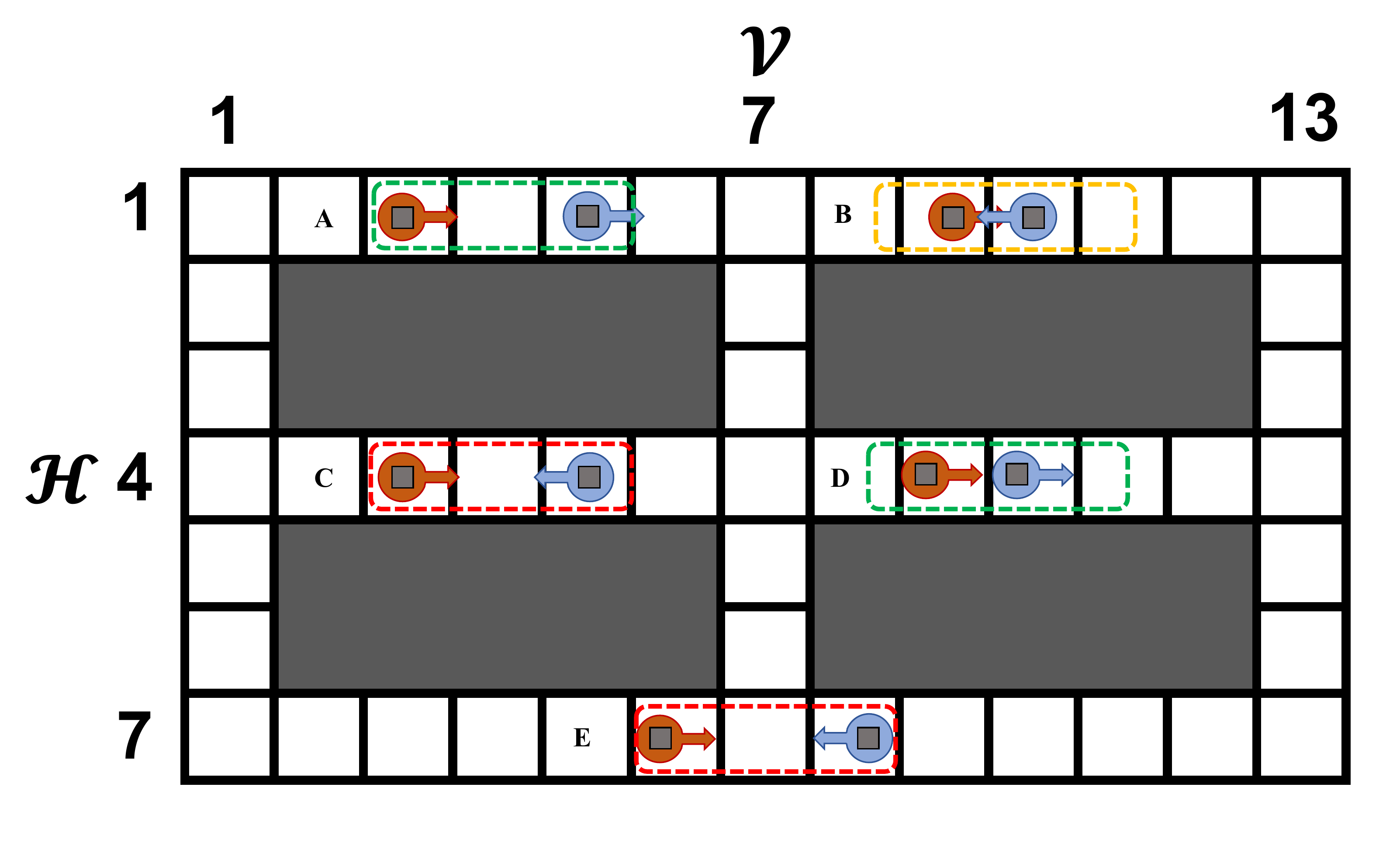}
    \caption{Amazon warehouse-like environment in a bird's eye. The horizontal and vertical passage intex sets are $\mathcal{H}=\{1,4,7\}\subset \{1,\dots,7\}$ and $\mathcal{V}=\{1,7,13\}\subset \{1,\dots,13\}$ with $H_s=3$ and $W_s=6$.}
    \label{warehouse}
\end{figure}

We consider an amazon warehouse-like grid environment which is represented by a binary matrix $M$ of dimension $h\times w$. We place $S$ shelves in the environment.
And we use $M(i,j)=1$ to represent that the grid cell $(i,j)$ is occupied. To maximum the storage capacity of the warehouse area, most of the cells are occupied. In order to allow the robots to move with the warehouse, some cells are left empty as passage. 
In Fig.~\ref{warehouse}, we define the horizontal and vertical passage index sets $\mathcal{H}$ and $\mathcal{V}$ as $\mathcal{H} := \{i_1, i_2, \dots\} \subset \{1,2,\dots,h\}$ and $\mathcal{V} := \{j_1, j_2, \dots\} \subset \{1,2,\dots,w\}$. 
Here both $\mathcal{H}$ and $\mathcal{V}$ are ordered sets and neither $\mathcal{H}$ nor $\mathcal{V}$ is empty. We let $M(i,j)=0$ for all $(i,j)\in \mathcal{H}\times \mathcal{V}$. For simplicity, the size of each shelf block is fixed as $h_s\times w_s$ in the following discussion. We choose $h_s=2$ and $i_1=1, j_1=1$ to allow the robots to approach every shelf. 
And we choose $w_s\in\{1,2,\dots,w-2\}$ such that 
all $j_m,j_{m+1}\in \mathcal{V}, j_{m+1} - j_m=w_s+1 =: W_s$
and for all $i_n,i_{n+1}\in \mathcal{H}, i_{n+1}-i_n = h_s+1 = 3 =: H_s$.
In this case, the width of passage in this environment is 1 so we call it a narrow-passage environment.

\subsection{One-way constraint}
In this work, we focus on the narrow-passage amazon warehouse-like environment above. Based on the assumptions above, in order to tackle the MPP problem in narrow-passage environment, we propose the particular lemma as follows.

\begin{lemma}
\label{lemma1}
\textbf{One-way constraint} Under Assumption~\ref{assum1}, to obtain an anytime-feasible path set $P$ on $G$, it requires that all possible robots pass each edge $e\in E$ in a unique direction. That is, for all $r\in R, t^0_r\in \mathbb{N},T_r\in \mathbb{N}$, if edge $\left(p_r(t),p_r(t+1)\right)$ exists in any $P_r$, the reversed edge $\left(p_r(t+1),p_r(t)\right)$ cannot exist in any other individual path of $P$.
\end{lemma}

\begin{Proof}

By contradiction, if $\big(p_r(t+1),p_r(t)\big)$ also exists in another path $P_s$, then by changing $t_s^0$, then it is  always possible to generate collision between $P_r$ and $P_s$.\hfill~$\blacksquare$
\end{Proof}

If robots collide in the crossings of passages, simple waiting strategy can be used to resolve the collision, i.e., robot near one crossing should wait until no robot in this crossing. Here, time optimality TTMPP (Total Time MPP) and MTMPP (Maximum Time MPP) can be achieved directly.

\section{General Methodology}
In this work, our methodology for multi-robot path planning problems can be divided generally into a three-step process, which is outlined in Algorithm~\ref{pipeline}. The proposed MPP problems are NP-hard, we introduce specific methods to simplify the problems. The key idea of methodology is constructing IP model based on the one-way constraint, thus, we call the methodology as \textbf{OWIP} (One-way Integer Programming).

\begin{figure}
    \centering
    \includegraphics[width=\linewidth]{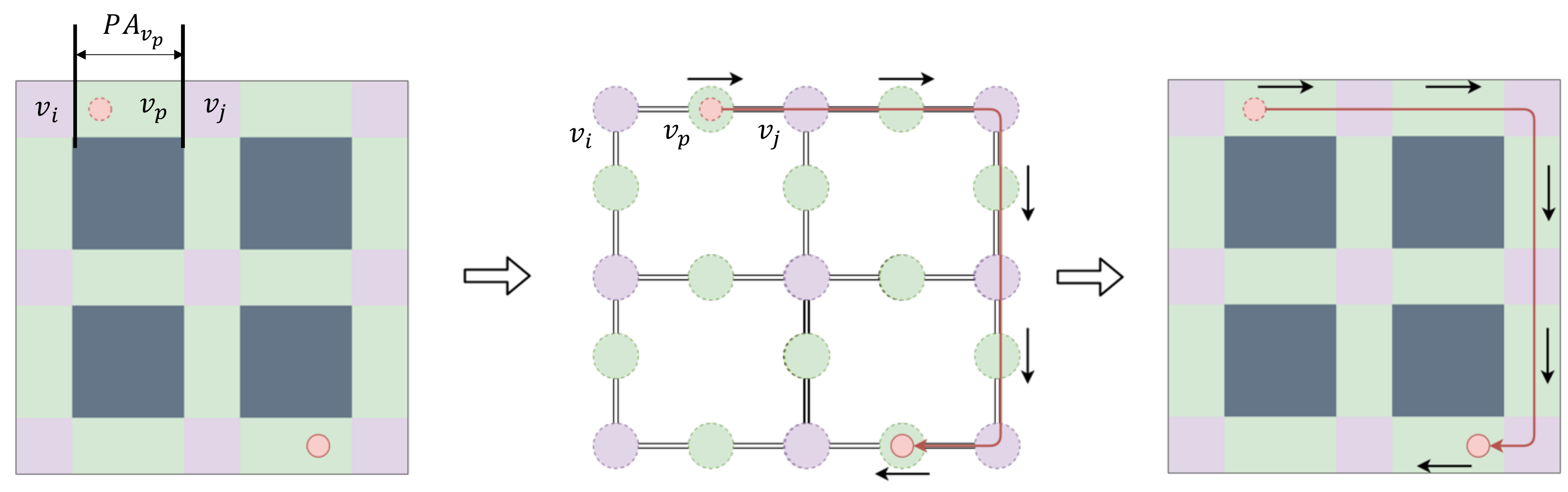}
    \caption{Complete algorithm process to get path set $P$ (other robots are omitted for clearness).}
    \label{project}
\end{figure}
\begingroup
\removelatexerror
\begin{algorithm}[!htpb]
  \caption{The \textbf{OWIP} process} 
  \label{pipeline}
  \small{
  \KwIn{$X^I,X^G,G$
  }
  \KwOut{$P$}
  \textbf{Topological extraction}: $\bar{X}^I, \bar{X}^G,\bG \leftarrow \mathrm{Extract}\left(X^I,X^G,G\right)$

  \textbf{IP model formulation}: $\bP \leftarrow \mathrm{IPsolver}\left(\bX^I,\bX^G,\bG\right)$
  
  \textbf{Projection}: $P \leftarrow \mathrm{Project}(\bP,X^I,X^G,G)$
  }
\end{algorithm}
\endgroup

\subsection{Topological Extraction}

As a direct result of Lemma~\ref{lemma1}, all the robots should pass each narrow passage in one fixed direction, which we call the \emph{one-way constraint}, and collisions in narrow passages can be avoided. Under such case, robots' motions in one narrow passage can be
determined by two vertices $v_i, v_j$ at both ends and one vertex $v_p$ in the passage. Inspired by \cite{nguyen2017generalized}, each passage can be compressed into an meta vertex along with two vertices in both ends of passages, which can significantly simplify the path planning. 

Here, as we denote in Fig.~\ref{project}, we call $v_i, v_j$ as crossing vertices and $v_p$ as passage vertex, the passage through $v_p$ can be denoted as $PA(v_p)$, whose length is $|PA(v_p)|$. A single passage vertex $v_p\in V_p$ is set for each individual passage, $V_p$ is the set of passage vertex, a crossing vertex $v_c\in V_c$ is the set for each crossing of passages in $G$, $V_c$ is the set of crossing vertex. In the first subfigure of Fig.~\ref{project}, green squares represent the passage vertices, pink squares represent the crossing vertices of passages in $G$.  In the second subfigure, green circles represent the passage vertices, pink circles represent the crossing vertices in $\bG$, black arrow indicates the one-way direction of passage. In the last subfigure, final path is achieved.

Therefore, with the similar definition of $G=(V,E)$, we introduce the \emph{topological map}:

\begin{definition}
The \emph{topological map} associated with $G=(V,E)$ is defined as: $ \bG=(\bV,\bE)$ whose vertex set is $\bV:= V_c\cup V_p$.
\end{definition}

For all vertices in $X^I$ and $X^G$, we find their corresponding nearest vertices in $\bV$: 

For each $r\in R$, if $X^I(r)\in V_c$ $\left(X^G(r)\in V_c\right)$, $\bX^I(r)=X^I(r)$ $\left(\bX^G(r)=X^G(r)\right)$; else $\bX^I(r)=\arg\min_{v\in V_p}|v-X^I(r)|$ $\left(\bX^G(r)=\arg\min_{v\in V_p}|v-X^G(r)|\right)$. 

Here, $|v_i-v_j|$ denotes the Manhattan distance between two vertices. Note that all the proposed $G=(V,E)$ can be extracted into $\bG=(\bV,\bE)$ and  $X^I,X^G$ can be transformed to $\bX^I,\bX^G$ in low polynomial time, which is shown in Fig.~\ref{project}. As will be discussed, with $\bX^I,\bX^G$, the IP model will be formulated and solved in $\bG$ without loss of generality. Thus, the number of decision variables in proposed IP model can be eliminated significantly.


\subsection{Topological-map-based IP model formulation}
    There are studies \cite{schouwenaars2001mixed,han2019integer,yu2016optimal} showing that the path planning problem can be rewritten as a linear program with integer constraints that count for the path continuity and collision avoidance. A key benefit of this IP approach is that the path planning optimization can be readily solved using the GUROBI optimization software with an MATLAB interface.
    
    \textbf{(Path continuity constraints)}  On topological map $\bG$, a path set $\bP$ containing the scheduled non-cyclic paths for all the robots: $\forall t^0_r\leq i< j\leq t^0_r+T_r, p_r(i)\neq p_r(j)$. Given $\langle \bG,\bX^I,\bX^G\rangle$, the topological-map-based IP model introduces a 3-D binary variable $x_{v_i,v_j,r}$ for each edge $(v_i,v_j)\in \bE$ to indicate whether $\bP_r$ uses $(v_i,v_j)$, i.e., whether robot $r$ passes $(v_i,v_j)$. For all $r\in R$, the following constraints must be satisfied:
    \begin{equation}
    \label{SG1}
        \sum_{v_i\in \bV,v_i\neq \bX^I(r)}x_{\bX^I(r),v_i,r}=\sum_{v_i\in\bV,v_i\neq \bX^G(r)}x_{v_i,\bX^G(r),r}=1;
    \end{equation}
    
     \begin{equation}
     \label{SG2}
        \sum_{v_i\in \bV,v_i\neq \bX^I(r)}x_{v_i,\bX^I(r),r}=\sum_{v_i\in\bV,v_i\neq \bX^G(r)}x_{\bX^G(r),v_i,r}=0;
    \end{equation}
    
     \begin{equation}
     \label{continue}
        \forall v_i\in \bV\backslash \{\bX^I(r),\bX^G(r)\},
        \sum_{v_j\in \bV}x_{v_i,v_j,r}=\sum_{v_j\in\bV}x_{v_j,v_i,r}\leq 1;
    \end{equation}
    
    Here, constraint~(\ref{SG1}) and (\ref{SG2}) make $\bP_r$ starts from $\bX^I(r)$ and terminates at $\bX^G(r)$. Constraint~(\ref{continue}) ensures that for each vertex, one outgoing edge can be used if and only if an incoming edge is used. The constraints above ensure the continuity of $\bP_r$ and forces each vertex to appear in $\bP_r$ at most once.
    
    As illustrated in \cite{miller1960integer}, a solution from IP model with constraint~(\ref{SG1}), (\ref{SG2}) and (\ref{continue}) could contain \emph{subtours}, which are cycles formed by edges that are disjoint from $\bP_r$. \cite{miller1960integer} denotes that subtours can be eliminated by creating integer decision variables $3\leq u_{v_i,r}\leq |\bV|$ for each $v_i\in \bV\backslash \{\bX^I(r),\bX^G(r)\}$, thus, new constraint should be added into the proposed IP model:
    \begin{equation}
        u_{v_i,r}-u_{v_j,r}+|\bV|x_{v_i,v_j,r}\leq |\bV|-1;
    \end{equation}
    
    \begin{proposition}
    \label{sol}
    Given $\langle \bG,\bX^I,\bX^G\rangle$, there exists a bijection between solutions to the topological-map-based IP model and all non-cyclic paths in $\bG$ from $\bX^I$ to $\bX^G$.
    \end{proposition}
    
    \begin{Proof}
    \emph{Injectivity.} Assume that adjacent matrix $x$ representing all the edge variables is a feasible solution of the topological-map-based IP model, a non-cyclic path starts from $\bX^I(r)$ to $\bX^G(r)$ can be achieved by following the positive edge variables in $x$ until reaching $\bX^G(r)$.
    
    \emph{Surjectivity.} Assume that $\bP_r=\{p_r(t^0_r),\dots,p_r(t^0_r+T_r)\}$ is feasible, $\bP_r$ can be converted into a feasible solution to the proposed IP model by assigning the corresponding edge variable in adjacent matrix $x$ to 1 and others to 0. \hfill~$\blacksquare$
    \end{Proof}
    
    After topological extraction, there could exist some $r$, $\bX^I(r)=\bX^G(r)$, we split $\bX^I(r)(\bX^G(r))$ into two vertices $v_{in},v_{out}\in V_c$ depending to the relative position of $X^I(r)$ and $X^G(r)$: $\bX^I(r)=v_{in},\bX^G(r)=v_{out}$.

    \textbf{(One-way constraints)} As we denote in Lemma~\ref{lemma1}, for each narrow-passage in $G$, one-way constraint should be imposed to prevent the robots from causing collisions within the passages.
    \begin{enumerate}
        \item if neither $\bX^I(r)$ nor $\bX^G(r)$ is in $PA_{v_p}$, the direction of robot $r$ in $PA_{v_p}$ can be represented as: 
        \begin{equation}
        \label{dir1}
            d(v_p,r)=x_{v_i,v_p,r}+3x_{v_p,v_i,r};
        \end{equation}
        
        \item if $\bX^I(r)$ is in $PA_{v_p}$, the direction of robot $r$ in $PA_{v_p}$ can be represented as: 
        \begin{equation}
        \label{dir2}
            d(v_p,r)=x_{v_p,v_j,r}+3x_{v_p,v_i,r};
        \end{equation}
        
        \item if $\bX^G(r)$ is in $PA_{v_p}$, the direction of robot $r$ in $PA_{v_p}$ can be represented as: 
        \begin{equation}
        \label{dir3}
            d(v_p,r)=x_{v_i,v_p,r}+3x_{v_j,v_p,r};
        \end{equation}
    \end{enumerate}
    
    \begin{proposition}
    Equation~\ref{dir1} to \ref{dir3} are able to denote all the directions of motion of all robots.
    \end{proposition}
    
    \begin{Proof}
        With constraint~\ref{SG1} to \ref{continue}, $P_r$ is continue and non-cyclic which is composed of vertices that do not repeat. Equation~\ref{dir1} ensures that robot $r$ which starts from $v_j(v_i)$, then passes passage $PA_{v_p}$ and terminates at $v_i(v_j)$, whose direction in $PA_{v_p}$ can be denoted with $x_{v_i,v_p,r}=x_{v_p,v_j,r},x_{v_j,v_p,r}=x_{v_p,v_i,r}$.
        Equation~\ref{dir2} and \ref{dir3} ensure that direction of robot $r$ can be denoted whose initial (goal) configuration is in $PA_{v_p}$. \hfill~$\blacksquare$
    \end{Proof}
    In general, the robot's direction in one passage can be $3$, $1$
 and $0$ (the robot does not pass this passage). Direction $3$ and Direction $1$ are the opposed. Thus, the general direction of passage $PA_{v_p}$ can be represented as:
 \begin{equation}
 \label{dir}
     D(v_p)=\max_{r\in R}d(v_j,r).
 \end{equation}
    With equation~\ref{dir}, similar to robot's direction, the direction of passage can be $3$, $1$
 and $0$ (no robot passes this passage). Direction $3$ and Direction $1$ are the opposed. Thus, one-way constraint can be translated into propose IP model, for all the passages $PA_{v_p}$:
 \begin{equation}
 \label{OWcons}
     \forall r\in R,\quad D(v_p)-d(v_p,r)\neq 2.
 \end{equation}
 
 Now, all the constraints in topological-map-based IP model in $\bG$ have been formulated with constraint~\ref{SG1} to \ref{OWcons}.
 
 \textbf{(Objective functions)} In this work,  we work on two common objectives focusing on distance optimality: \emph{Total Distance} and \emph{Maximum Distance}. Now we translate them into IP model objective functions.
 
 In topological map $\bG$, for each edge $(v_i,v_p)\in \bE$, we assume that $v_p$ is the passage vertex. Thus edge $(v_i,v_p)$ can be related to certain weight which is decided by the length of passage $|PA_{v_p}|$ (see Fig.~\ref{project}). Denote the correlated weights as:
 \begin{equation}
     \{w_{v_i,v_p}=w_{v_p,v_i}=|v_p-v_i|\; | (v_i,v_p),(v_p,v_i)\in \bE\}.
 \end{equation}
 
 \textbf{Minimize Total Distance:} For all $r\in R$, with binary decision variable $x_{v_i,v_j,r}$ for each edge $(v_i,v_j)\in \bE$ to indicate whether $P_r$ uses $(v_i,v_j)$, the objective function of \emph{total distance} can be denoted as:
 \begin{equation}
     \sum_{r\in R}\sum_{(v_i,v_j)\in \bE}w_{v_i,v_j}x_{v_i,v_j,r};
 \end{equation}
 
 \textbf{Minimize Max Distance: } To denote the minimize max distance objective function, we introduce an additional integer decision variable $xmax$ and for all $r\in R$ add the constraint:
 \begin{equation}
 \label{minmax}
     \sum_{(v_i,v_j)\in \bE}w_{v_i,v_j}x_{v_i,v_j,r}\leq xmax(r).
 \end{equation}
 
 In constraint~\ref{minmax}, the left side represents the distance traveled by robot $r$. Thus, the objective function can be denoted as:
 \begin{equation}
     \min_{r\in R} xmax(r).
 \end{equation}
 
 As the collision avoidance in narrow passages is ensured by one-way constraint, the improvement of time optimality can be then considered by reducing the likelihood that the robot will wait at the intersection vertex. Here, new data structure $\mathcal{P}$ of dimension $n\times |V_c|$ is proposed to record the vertices $v\in V_c$ which are passed by robot $r$.
 
 \begin{equation}
 \label{number}
     \mathcal{P}(r,v)=\max_{v_i\in V_c}{x_{v,v_i,r}}.
 \end{equation}
 
 Then we can denote the number of robots passing at the intersection vertex $v\in V_c$:
 
  \begin{equation}
 \label{cost}
     C(v)=\sum_{r\in R}\mathcal{P}(r,v).
 \end{equation}
 In order to reduce the congestion in each intersection vertex, we can add $\max_{v\in V_c }{C(v)}$ to each distance optimality objective function to improve the time optimality.
 Now, with complete constraints and two objective functions, topological-map-based IP model has been formulated completely. 

Here, suppose that scheduled path set $\bP$ is the solution of the proposed IP model with proposition~\ref{sol}. We note that for all $v_i\in \bP_r$, $v_i\in \bV$, in order to retain the scheduled path set $P$ in $G$, vertices in $V$ should be added between $v_i$ and $v_j$, $v_i,v_j\in \bP_r, (v_i,v_j)\in\bE$. As the proposed IP model ensures that the passages being passed by robots are one-way, the initial and goal configuration $X^I$ and $X^G$ can be added into the scheduled path set directly. For path set $P$, we ensure that for all $t_r^0\leq t < t_r^0+T_r$:
\begin{equation}
    |p_r(t+1)-p_r(t)|=1, p_r(t),p_r(t+1)\in V.
\end{equation}

The complete algorithm process can be seen in Fig.~\ref{project} visually. With the three-step process of \textbf{OWIP}, we have the proposition :
\begin{proposition}
    Algorithm \textbf{OWIP} is a complete algorithm for finding MPP solution in narrow-passage warehouse-like environment that minimize the total distance or minimize the max distance traveled by robots.
\end{proposition}

\section{Heuristic feasible initial solution}
Intuitively, the time to solve an IP model is often influenced by the number of decision variables in the model, by topological extraction, the number of decision variables have been reduced significantly. In order to further increase the efficiency of solving the IP model, we consider that in cases where the IP solver is slow in finding an initial feasible solution, it can be helpful to provide a feasible solution along with the model itself. Thus, we propose a heuristic algorithm based on $\bG$ to find the anytime-feasible initial solution in polynomial running time that satisfy the path continuity constraints and one-way constraints.




\begin{figure}
    \centering
    \includegraphics[width=\linewidth]{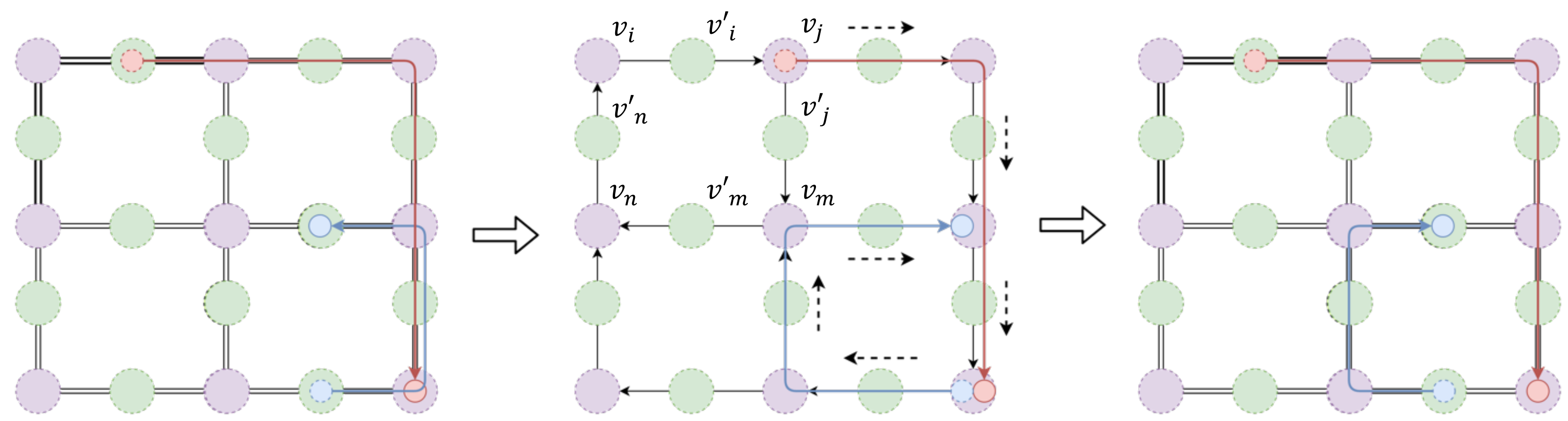}
    \caption{[Left] Topological map with collision-ignorance path set of robots. [Middle] Topological map with one-way directions (indicated by black arrows). [Right] Topological map with initial feasible solutions.}
    \label{heu}
\end{figure}
\subsection{Overview}
The proposed heuristic algorithm adopts a four-step process. The key idea is that we base on the IP model input $\bX^I,\bX^G, \bG$ to retain initial any-time feasible solution $\bP^0$, then provide $\bP^0$ along with the model itself to Gurobi solver.

\begingroup
\removelatexerror
\begin{algorithm}[!htpb]
  \caption{The heuristic algorithm pipeline} 
  \label{overview}
  \small{
  \KwIn{$\bX^I,\bX^G,\bG$
  }
  \KwOut{$\bP^0$}
  \textbf{Projection}: $x^I, x^G \leftarrow \mathrm{Projection}\left(\bX^I,\bX^G,\bG \right)$

  \textbf{Initial path planning}: $\mathcal{O},\mathcal{N} \leftarrow \mathrm{InitialPathPlan}(x^I,x^G,\bG)$
  
  \textbf{One-way regulation}: $\bG \leftarrow \mathrm{OneWayRegulation}(\mathcal{O},\mathcal{N},\bG)$

  \textbf{Final path planning}: $\bP^0\leftarrow \mathrm{FinalPathPlan}\left(\bX^I,\bX^G,\bG\right)$
  }
\end{algorithm}
\endgroup

\subsection{Projection}

A projection from $\bX^I$ and $\bX^G$ in $V_p$ to $V_c$ is implemented in Algorithm~\ref{projectionPipe}. It is possible that the two crossing vertices in both ends of narrow passage have the same Manhattan distance from $\bX^I(r)$ or $\bX^G(r)$. 
In such case, we can choose an arbitrary one from them.

\begingroup
\removelatexerror
\begin{algorithm}[!htpb]
  \caption{Projection$(\bX^I,\bX^G,\bG)$} 
  \label{projectionPipe}
  \small{
  \KwIn{$\bX^I$, $\bX^G$, $\bG$
  }
  \KwOut{$x^I$, $x^G$}
  \setcounter{AlgoLine}{0}

  \ForAll(){
    $r \in R$
  }{
    $x^I(r) = \mathop{\arg\min}_{v\in V_c} |\bX^I(r) - v| $ 
      
    $x^G(r) = \mathop{\arg\min}_{v\in V_c} |\bX^G(r) - v| $      	
  }
  
  \Return{$x^I,x^G$}
  }
\end{algorithm}
\endgroup

\subsection{Initial path planning}

Here, collision-ignore path set $\bP^0$ is found with A* search while ignoring other robots on $\bG$. We note that the edge set associated with path $P_r$ on $\bG$ can be defined as: $\mathcal{E}(P_r):=\left\{(P_r(t),P_r(t+1))|\forall 0\leq t < len(P_r), \left(P_r(t),P_r(t+1)\right) \in \bE\right\}$.

We use an integer array $\mathcal{O}:|\bE|\times 1$ which stores the sum of the distance spent by all the robots on each edge of $\bE$. $\mathcal{N}:|\bE|\times |R|$ is a binary matrix which indicates the edges traveled by each robot.

\begingroup
\removelatexerror
\begin{algorithm}[!htpb]
  \caption{InitialPathPlan$(\bar{X}^I,\bar{X}^G,\bG$)} 
  \label{InPath}
  \setcounter{AlgoLine}{0}
  \small{
  \KwIn{$\bar{X}^I,\bar{X}^G,\bG$
  }
  \KwOut{$\mathcal{O},\mathcal{N}$}
  
  $\mathcal{O}(\bE)=\mathbf{0}$
  
  \ForAll()
  {
  $r\in R$
  }
  {
    $\bP^0_r=A^*\left(\bG,x^I(r),x^G(r)\right)$

    \ForAll(){
      $e\in \mE({\bP^0_r})$
    }{
      $\mathcal{O}(e)=\mathcal{O}(e)+|e|$ 
      
      $\mathcal{N}(e,r)=1$
    }
  }
  
  \Return{$\mathcal{O},\mathcal{N}$}
  
  }
\end{algorithm}
\endgroup

\subsection{One-way regulation}

In the previous step, the initial path for each robot is achieved ignoring other robots. Now, we introduce an mechanism \emph{One-way Regulation} based on the one-way constraint to resolve collision. First, we define a topological gadget \emph{0-instance} whose schematic diagram is given in the middle part of Fig.~\ref{heu}. In our proposed narrow-passage warehouse-like environment, each shelf can be surrounded by four crossing vertices: $v_i,v_j,v_m,v_n \in V_c$, and four passage vertices: $v'_i,v'_j,v'_m,v'_n$.

\begin{definition}
    A \emph{0-instance} on $g$ is defined as a vertex loop on $g$ surrounding each shelf $s$, $s\in \{1,\dots,S\}$. We define the following \emph{clockwise 0-instance} as a vertex loop set of adjacent vertices:
    $I_s:=\{v_i,v'_i,v_j,v'_j,v_m,v'_m,v_n,v'_n,v_i\}$ with \emph{clockwise 0-instance} $I'_s:=\{v_i,v'_n,v_n,v'_m,v_m,v'_j,v_j,v'_i,v_i\}$.
\end{definition}

Each vertex set $V_s$ has a corresponding edge set: $\mathcal{E}(V_s):=\{\left(V_s(i),V_s(i+1)\right)|\forall 1\leq i<|I_s|\}$ We introduce an \emph{Instance Rotation} operation to switch a clockwise (counterclockwise) 0-instance to (counterclockwise) clockwise 0-instance.  Now $\bG$ is undirected, so that for any $s$, $I_s$ and $I'_s$ simultaneously exist, i.e., $\mE(I_s)\in \bE, \mE(I'_s)\in \bE$.

Because of one-way constraint, for any $s$, one and only one of $I_s$ and $I'_s$ can exist. In order to choose the better one between $I_s$ and $I'_s$, heuristic value for $I_s$ and $I'_s$ should be calculated. 

Here, we propose a heuristic value $H_v(s)$ which represents the total detour cost: in order to satisfy the one-way constraint, some robots cannot follow the collision-ignore paths to move (see Fig~\ref{project} for example). Thus, detour costs in scheduled paths are caused with respect to collision-ignore paths.

Total detour cost $D_t$ for $I_s$ and heuristic value $H_v(s)$ are proposed as:

\begin{equation}
\label{detour}
\begin{aligned}
        D_t(I_s)=\left(\sum\nolimits_{r\in \{1,2\dots,R\}}
        \min\left(
            \sum\nolimits_{e\in \mE({I_s})}\mathcal{N}(e,r),1
            \right)
    \right)\\
    \cdot\sum\nolimits_{e\in \mE(I_s)}|e|-\sum\nolimits_{e\in \mE({I_s})}{\mathcal{O}(e).}
\end{aligned}
\end{equation}

\begin{algorithm}[!htpb]
  \caption{DetourCost$(\mathcal{N},\mathcal{O},\bG)$} 
  \label{DC}
  \setcounter{AlgoLine}{0}
  \small{
  \KwIn{$\mathcal{N},\mathcal{O},\bG$
  }
  \KwOut{$\bG$}
  
  \ForAll(){$s\in\{1,\dots,S\}$}{
  
  $H_v(s)=\max{(D_t(I_s),D_t(I'_s)}$ \tcp{According to equation~\ref{detour}}
  
  $I_s=\arg\max_{I\in\{I_s,I'_s\}}{(D_t(I))}$ 
  
  }
   
  Sort all $I_s, s\in\{1,\dots,S\}$ in the descending order of $H_v(s)$.
  
    $E_n=\mE(I_1),V_n=I_1$
    
    \ForAll(){$s\in \{2,\dots,S\}$}{

      $E_n=E_n+\mE(I_s)-\mE(V_n\cap I_s)$
      
        $V_n=V_n+I_s$
            
      }
    $\bV=V_n,\bE=E_n$
  }
  
  \Return{$\bG$}
\end{algorithm}

The \emph{One-way Regulation} process is denoted in Algorithm.~\ref{DC}. After one-way regulation, we know that $\bG$ is directed: if $(v_i,v_j)\in \bE$, then $(v_j,v_i)\notin \bE$ and vice versa.

\begin{theorem}
    \label{FeasibleMotion}
    Under one-way constraint, $\bG$ satisfies that any $v_s,v_t\in \bV$, there exists at least one path from $v_s$ to $v_t$ if each $\mE({I_s})$ is regulated by \emph{Instance Rotation}. 
\end{theorem}

\begin{Proof}
    It is no doubt that for all $s\in \{1,2,\dots,S\}$, $v_1,v_2\in I_s$, there exists a path from $v_1$ to $v_2$ as all the edges in $\mE({I_s})$ are connected end to end. Furthermore, $\forall e\in\bE,\exists s\in \{1,2,\dots,S\},e\in \mE({I_s})$ can be confirmed with Algorithm~\ref{DC}. Thus, theorem~\ref{FeasibleMotion} can be satisfied.\hfill$\blacksquare$
\end{Proof}

\begin{figure}[ht]
    \centering
    \includegraphics[width=1\linewidth]{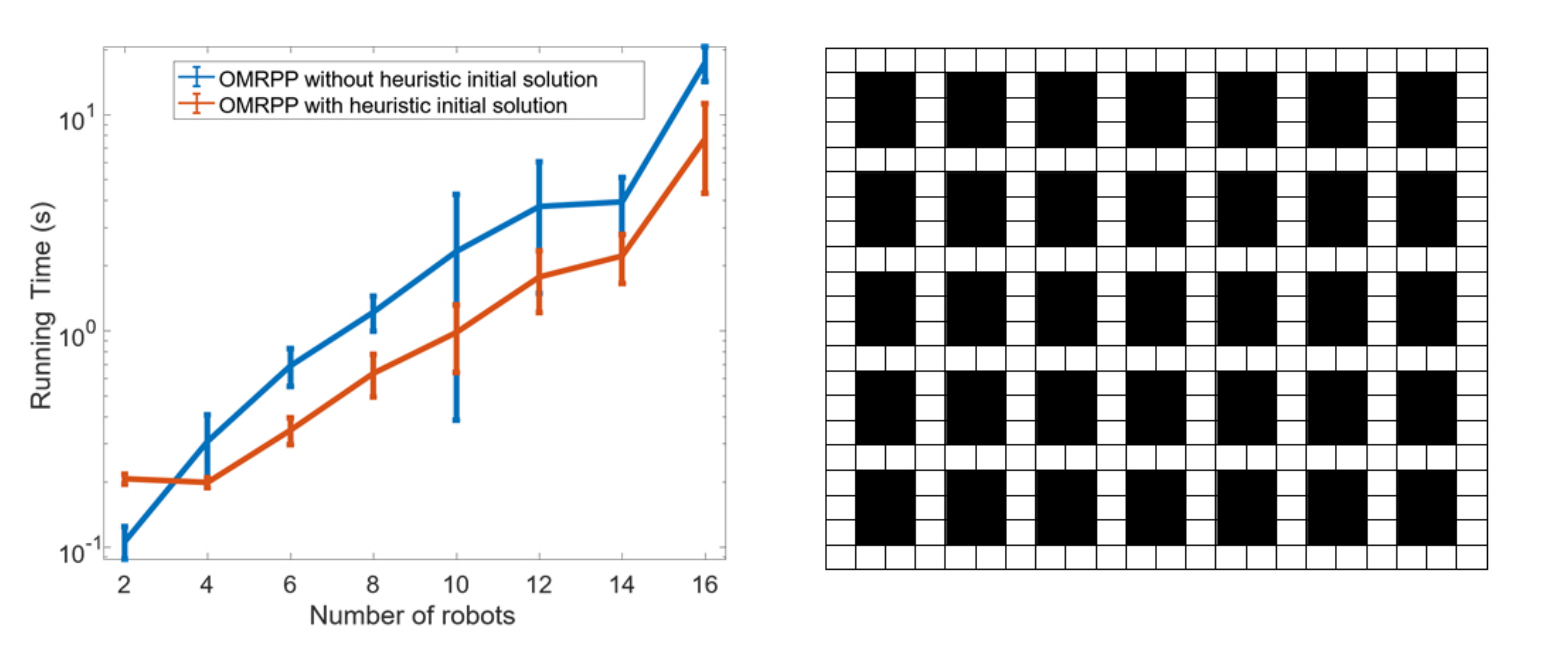}
    \caption{[Left]Running time comparison for heuristic feasible initial solution. [Right]Simulation environment.}
    \label{cmp}
\end{figure} 

The left part of Figure.~\ref{cmp} proves the effectiveness of heuristic feasible initial solution to reduce the running time of OMRPP.

\subsection{Final path planning}

Here, the one-way passages are formed on the directed $\bG$.
Feasible initial solution $\bP^0$ is found by applying heuristic $A^*$ algorithm. 
With proposition~\ref{sol}, initial solutions value for decision variables $x$ can be assigned by $\bP^0$.

\begin{proposition}
    Time complexity of heuristic feasible initial solution algorithm is polynomial ($O(n|\bV|^2)$).
\end{proposition}

\begin{Proof}
    In Algorithm~\ref{overview}, projection (Line 1)
has time complexity of $O\left(n \right)$.
In Line 2, we apply the heuristic $A^*$ search which has time complexity \footnote{We reduce the time complexity of $A^*$ search to the time complexity of Dijkstra search by improving the operations on $OPEN$ list. Explanation is omitted here due to space limitation.} of  $O(n|\bV|^2)$ . 
In Line 3, \emph{One-way regulation} can be terminated in $O(n|\bV|^2+|\bV|^2)$.
In Line 4 the final paths can be searched in $O(n|\bV|^2)$ with heuristic $A^*$ Algorithm.
Thus, \textbf{OMRPP} has time complexity of $O(n|\bV|^2)$.
\hfill$\blacksquare$
\end{Proof}

\section{Experiments and results}
\begin{figure*}[ht]
    \centering
    \includegraphics[width=1\linewidth]{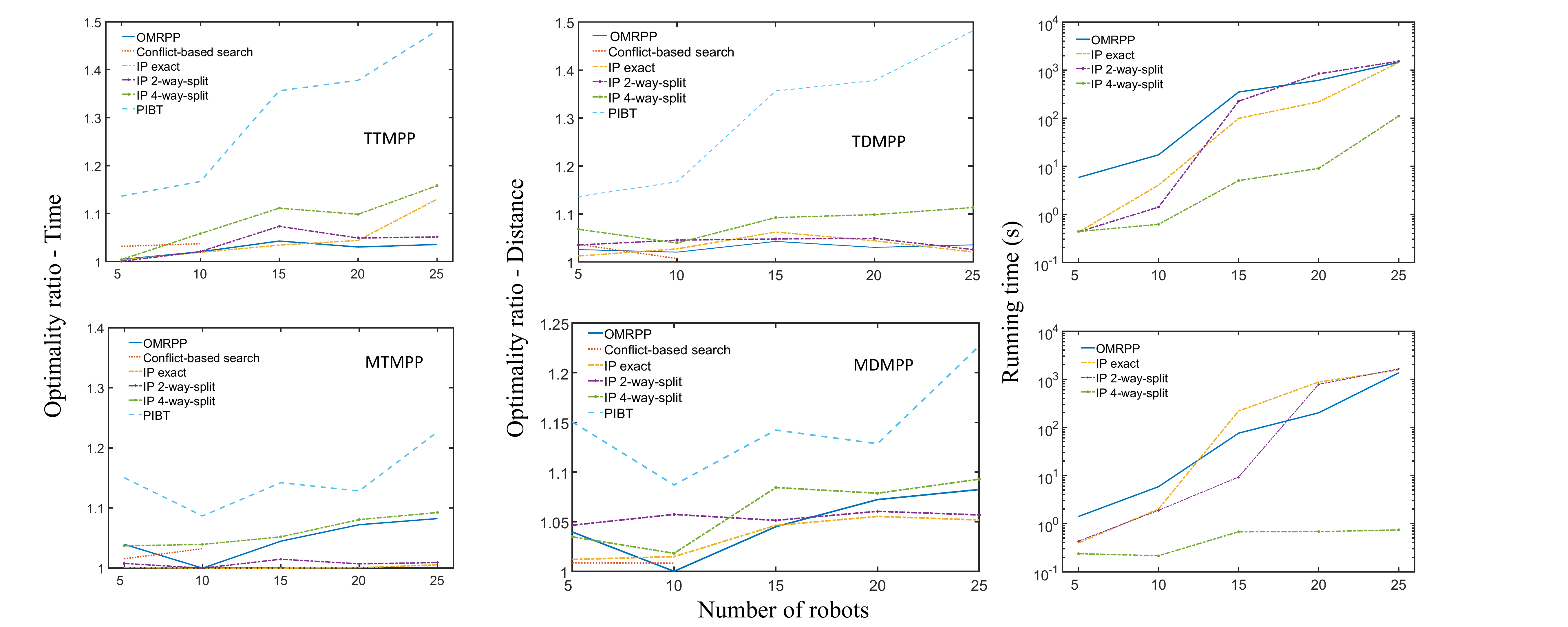}
    \caption{[Upper]Simulation results of total time and total distance optimality. [Lower]Simulation results of maximum time and maximum distance optimality.}
    \label{res}
\end{figure*}

In this work, we evaluate \textbf{OMRPP} on warehouse-like narrow-passage environment. Here, we represent part of results in this section. We construct a $21\times 22$ such grid environment (right part of Figure~\ref{cmp}) for simulation.

\subsection{Experiment setup}
We implement \textbf{OMRPP} and CBS in MATLAB. Concerning k-way split IP, we use the existing implementation by \cite{yu2016optimal} written in Java. We use PIBT method by \cite{okumura2019priority} written in C++.
Here we note that the constructed IP models in \textbf{OMRPP} are solved using Gurobi Solver (Academic Version 9.1.1). All experiments are executed on an Intel(R) Xeon(R) CPU E5-2650 v4 with 128GB RAM at 2.20GHz on Ubuntu 5.4.0. The experimental evaluation has been done on diverse instances consisting of robot set with random initial and goal configuration ranging in the number of robots from small to large. We also compare the running time of IP-based MPP algorithms, i.e., \textbf{OMRPP}, IP exact and IP k-way split $k=2,4$.

We vary the number of robots from 10 to 50 while initial and goal configurations of robots are generated randomly for each single robot in robot set.

\subsection{TDMPP and TTMPP problem}
Simulation results are represented in the upper part of Fig.~\ref{res}. We present \emph{optimality ratio} which is measured as the solution cost over an underestimated cost, i.e., the solution cost of the paths set ignoring the robot collision.

The general tend we can observe from Fig.~\ref{res} indicates that the performance of \textbf{OMRPP} degrades as number of robots gets bigger. Comparing with PIBT and IP 4-way split, \textbf{OMRPP} has better performance while CBS, IP exact and IP 2-way split achieve lower solution cost but longer running time (CBS cannot be solved in $10^4$ seconds with robots exceeding 20 robot).  For each number of robot, 10 times of test are done to achieve the mean value for optimality ratio and \emph{running time } for each number of robot, i.e., time to achieve the optimal solution of IP model.

Thus, we can conclude that in case of TDMPP and TTMPP problem, \textbf{OMRPP} has a relatively efficient performance when the number of robots increases.

\subsection{MDMPP and MTMPP problem}

In case of MDMPP and MTMPP problem, we can see the performance of \textbf{OMRPP} under the narrow-passage warehouse-like environment in the lower part of Fig.~\ref{res}. In comparison with PIBT, \textbf{OMRPP} always holds better optimality performance. Taking the other IP-based methods into consideration, when the number of robots grows, the optimality performance of \textbf{OMRPP} augments and exceeds IP 4-way split. Meanwhile, IP k-way split solves more quickly the problem with small number of robots than \textbf{OMRPP}, but the growth of \textbf{OMRPP} running time is much lower than IP 2-way split and IP exact, which can prove the efficiency of \textbf{OMRPP} under the narrow-passage warehouse-like environment.

\section{Conclusion}

In this work, we provide a fast and high-efficient algorithm to tackle the MPP problem in narrow-passage warehouse-like environments, which guarantees to generate collision-free paths. 
We tackle the bottleneck of MPP algorithms in narrow-passage environments with the proposal of anytime feasibility property based on one-way constraint. 
The proposed algorithm combines advantages of efficient running time and optimality performance. In future work, we would like to apply \textbf{OMRPP} heuristics for solving MPP problems beyond warehouse-like environment.




\bibliographystyle{./bibliography/IEEEtran}
\bibliography{./bibliography/IEEEabrv,./bibliography/IEEEexample}

\begin{thebibliography}{10}
\providecommand{\url}[1]{#1}
\csname url@samestyle\endcsname
\providecommand{\newblock}{\relax}
\providecommand{\bibinfo}[2]{#2}
\providecommand{\BIBentrySTDinterwordspacing}{\spaceskip=0pt\relax}
\providecommand{\BIBentryALTinterwordstretchfactor}{4}
\providecommand{\BIBentryALTinterwordspacing}{\spaceskip=\fontdimen2\font plus
\BIBentryALTinterwordstretchfactor\fontdimen3\font minus
  \fontdimen4\font\relax}
\providecommand{\BIBforeignlanguage}[2]{{%
\expandafter\ifx\csname l@#1\endcsname\relax
\typeout{** WARNING: IEEEtran.bst: No hyphenation pattern has been}%
\typeout{** loaded for the language `#1'. Using the pattern for}%
\typeout{** the default language instead.}%
\else
\language=\csname l@#1\endcsname
\fi
#2}}
\providecommand{\BIBdecl}{\relax}
\BIBdecl

\bibitem{yu2015intractability}
J.~Yu, ``Intractability of optimal multirobot path planning on planar graphs,''
  \emph{IEEE Robotics and Automation Letters}, vol.~1, no.~1, pp. 33--40, 2015.

\bibitem{han2020ddm}
S.~D. Han and J.~Yu, ``Ddm: Fast near-optimal multi-robot path planning using
  diversified-path and optimal sub-problem solution database heuristics,''
  \emph{IEEE Robotics and Automation Letters}, vol.~5, no.~2, pp. 1350--1357,
  2020.

\bibitem{wurman2008coordinating}
P.~R. Wurman, R.~D'Andrea, and M.~Mountz, ``Coordinating hundreds of
  cooperative, autonomous vehicles in warehouses,'' \emph{AI magazine},
  vol.~29, no.~1, pp. 9--9, 2008.

\bibitem{han2019effective}
S.~D. Han and J.~Yu, ``Effective heuristics for multi-robot path planning in
  warehouse environments,'' in \emph{2019 International Symposium on
  Multi-Robot and Multi-Agent Systems (MRS)}.\hskip 1em plus 0.5em minus
  0.4em\relax IEEE, 2019, pp. 10--12.

\bibitem{liu2019task}
M.~Liu, H.~Ma, J.~Li, and S.~Koenig, ``Task and path planning for multi-agent
  pickup and delivery,'' in \emph{Proceedings of the International Joint
  Conference on Autonomous Agents and Multiagent Systems (AAMAS)}, 2019.

\bibitem{fox2000probabilistic}
D.~Fox, W.~Burgard, H.~Kruppa, and S.~Thrun, ``A probabilistic approach to
  collaborative multi-robot localization,'' \emph{Autonomous robots}, vol.~8,
  no.~3, pp. 325--344, 2000.

\bibitem{luna2010network}
R.~Luna and K.~E. Bekris, ``Network-guided multi-robot path planning in
  discrete representations,'' in \emph{2010 IEEE/RSJ International Conference
  on Intelligent Robots and Systems}.\hskip 1em plus 0.5em minus 0.4em\relax
  IEEE, 2010, pp. 4596--4602.

\bibitem{erdem2013general}
E.~Erdem, D.~G. Kisa, U.~Oztok, and P.~Sch{\"u}ller, ``A general formal
  framework for pathfinding problems with multiple agents,'' in
  \emph{Twenty-Seventh AAAI Conference on Artificial Intelligence}, 2013.

\bibitem{erdmann1987multiple}
M.~Erdmann and T.~Lozano-Perez, ``On multiple moving objects,''
  \emph{Algorithmica}, vol.~2, no.~1, pp. 477--521, 1987.

\bibitem{lavalle1998optimal}
S.~M. LaValle and S.~A. Hutchinson, ``Optimal motion planning for multiple
  robots having independent goals,'' \emph{IEEE Transactions on Robotics and
  Automation}, vol.~14, no.~6, pp. 912--925, 1998.

\bibitem{schouwenaars2001mixed}
T.~Schouwenaars, B.~De~Moor, E.~Feron, and J.~How, ``Mixed integer programming
  for multi-vehicle path planning,'' in \emph{2001 European control conference
  (ECC)}.\hskip 1em plus 0.5em minus 0.4em\relax IEEE, 2001, pp. 2603--2608.

\bibitem{johnson1879notes}
W.~W. Johnson, W.~E. Story \emph{et~al.}, ``Notes on the “15” puzzle,''
  \emph{American Journal of Mathematics}, vol.~2, no.~4, pp. 397--404, 1879.

\bibitem{loyd1959mathematical}
S.~Loyd, \emph{Mathematical puzzles}.\hskip 1em plus 0.5em minus 0.4em\relax
  Courier Corporation, 1959, vol.~1.

\bibitem{archer1999modern}
A.~F. Archer, ``A modern treatment of the 15 puzzle,'' \emph{The American
  Mathematical Monthly}, vol. 106, no.~9, pp. 793--799, 1999.

\bibitem{goldreich2011finding}
O.~Goldreich, ``Finding the shortest move-sequence in the graph-generalized
  15-puzzle is np-hard,'' in \emph{Studies in complexity and cryptography.
  Miscellanea on the interplay between randomness and computation}.\hskip 1em
  plus 0.5em minus 0.4em\relax Springer, 2011, pp. 1--5.

\bibitem{van2008reciprocal}
J.~Van~den Berg, M.~Lin, and D.~Manocha, ``Reciprocal velocity obstacles for
  real-time multi-agent navigation,'' in \emph{2008 IEEE International
  Conference on Robotics and Automation}.\hskip 1em plus 0.5em minus
  0.4em\relax IEEE, 2008, pp. 1928--1935.

\bibitem{sharon2015conflict}
G.~Sharon, R.~Stern, A.~Felner, and N.~R. Sturtevant, ``Conflict-based search
  for optimal multi-agent pathfinding,'' \emph{Artificial Intelligence}, vol.
  219, pp. 40--66, 2015.

\bibitem{okumura2019priority}
K.~Okumura, M.~Machida, X.~D{\'e}fago, and Y.~Tamura, ``Priority inheritance
  with backtracking for iterative multi-agent path finding,'' \emph{arXiv
  preprint arXiv:1901.11282}, 2019.

\bibitem{de2013push}
B.~de~Wilde, A.~W. ter Mors, and C.~Witteveen, ``Push and rotate: cooperative
  multi-agent path planning,'' in \emph{Proceedings of the 2013 international
  conference on Autonomous agents and multi-agent systems}, 2013, pp. 87--94.

\bibitem{yu2016optimal}
J.~Yu and S.~M. LaValle, ``Optimal multirobot path planning on graphs: Complete
  algorithms and effective heuristics,'' \emph{IEEE Transactions on Robotics},
  vol.~32, no.~5, pp. 1163--1177, 2016.

\bibitem{han2019integer}
S.~D. Han and J.~Yu, ``Integer programming as a general solution methodology
  for path-based optimization in robotics: Principles, best practices, and
  applications,'' in \emph{2019 IEEE/RSJ International Conference on
  Intelligent Robots and Systems (IROS)}.\hskip 1em plus 0.5em minus
  0.4em\relax IEEE, 2019, pp. 1890--1897.

\bibitem{barer2014suboptimal}
M.~Barer, G.~Sharon, R.~Stern, and A.~Felner, ``Suboptimal variants of the
  conflict-based search algorithm for the multi-agent pathfinding problem,'' in
  \emph{Seventh Annual Symposium on Combinatorial Search}, 2014.

\bibitem{wang2008fast}
K.-H.~C. Wang, A.~Botea \emph{et~al.}, ``Fast and memory-efficient multi-agent
  pathfinding.'' in \emph{ICAPS}, 2008, pp. 380--387.

\bibitem{cohen2016improved}
L.~Cohen, T.~Uras, T.~S. Kumar, H.~Xu, N.~Ayanian, and S.~Koenig, ``Improved
  solvers for bounded-suboptimal multi-agent path finding.'' in \emph{IJCAI},
  2016, pp. 3067--3074.

\bibitem{nguyen2017generalized}
V.~Nguyen, P.~Obermeier, T.~C. Son, T.~Schaub, and W.~Yeoh, ``Generalized
  target assignment and path finding using answer set programming,'' in
  \emph{Proceedings of the 26th International Joint Conference on Artificial
  Intelligence}, 2017, pp. 1216--1223.

\bibitem{miller1960integer}
C.~E. Miller, A.~W. Tucker, and R.~A. Zemlin, ``Integer programming formulation
  of traveling salesman problems,'' \emph{Journal of the ACM (JACM)}, vol.~7,
  no.~4, pp. 326--329, 1960.

\end{thebibliography}

\vspace{12pt}

\end{document}